\documentclass{article}
\usepackage{ijcai25}

\usepackage{tcolorbox}
\usepackage{times}
\usepackage{soul}
\usepackage{url}
\usepackage[hidelinks]{hyperref}
\usepackage[utf8]{inputenc}
\usepackage[small]{caption}
\usepackage{graphicx}
\usepackage{amsmath}
\usepackage{amsthm}
\usepackage{booktabs}
\usepackage{algorithm}
\usepackage{algorithmic}
\usepackage[switch]{lineno}
\usepackage{xcolor}
\usepackage{subcaption}

\usepackage{multirow} % Required for multirows
\usepackage{amsmath,amssymb,amsfonts}
\usepackage{algorithmic}
\usepackage{textcomp}
\usepackage{booktabs}
\usepackage{listings}
\colorlet{punct}{red!60!black}
\colorlet{numb}{magenta!60!black}
\definecolor{background}{HTML}{EEEEEE}
\definecolor{delim}{RGB}{20,105,176}

\lstdefinelanguage{Json}{
    showstringspaces=false,
    breaklines=true,
    frame=lines,
    backgroundcolor=\color{background},
    literate=
     *{0}{{{\color{numb}0}}}{1}
      {1}{{{\color{numb}1}}}{1}
      {2}{{{\color{numb}2}}}{1}
      {3}{{{\color{numb}3}}}{1}
      {4}{{{\color{numb}4}}}{1}
      {5}{{{\color{numb}5}}}{1}
      {6}{{{\color{numb}6}}}{1}
      {7}{{{\color{numb}7}}}{1}
      {8}{{{\color{numb}8}}}{1}
      {9}{{{\color{numb}9}}}{1}
      {:}{{{\color{punct}{:}}}}{1}
      {,}{{{\color{punct}{,}}}}{1}
      {\{}{{{\color{delim}{\{}}}}{1}
      {\}}{{{\color{delim}{\}}}}}{1}
      {[}{{{\color{delim}{[}}}}{1}
      {]}{{{\color{delim}{]}}}}{1},
}

\newtcolorbox{mybox}[2][]{
  colback=gray!10!white,
  colframe=black!75!black,
  fonttitle=\bfseries,
  coltitle=white,
  rounded corners,
  boxrule=0.5mm,
  title=#2,#1,
}

\title{Tug-of-War within A Decade: Conflict Resolution in Vulnerability Analysis via Teacher-Guided Retrieval-Augmented Generations}

\author{
Ziyin Zhou$^1$
\and
Jianyi Zhang$^1$*\and
Xu ji$^{1}$\and
Yilong Li$^1$\and 
Jiameng Han$^1$\and
Zhangchi Zhao$^1$
\affiliations
$^1$Beijing Electronic Science and Technology Institute\\
\emails
zjy@besti.edu.cn
}

\begin{document}

\maketitle

\begin{abstract}
% 2025.01.18改了第二遍
% 大型语言模型（llm）对于分析和解决网络安全漏洞至关重要。然而，在过去十年中发现的20多万个漏洞中，有3万多个已被更改或更新。这就需要经常更新训练数据集和LLM的内部知识库，以保持知识的一致性。由于训练周期的限制，模型内部知识的不一致性随着时间的推移而加剧，使得检测新出现的威胁变得越来越复杂。本文主要针对CVE （Common Vulnerabilities and Exposures）检测与分析内容中存在的知识差异和冲突问题进行研究。这个问题阻碍了LLM从原始训练数据集中检索最新知识的能力，导致知识冲突，捏造事实不正确的结果，并产生幻觉。为了解决这个问题，我们提出了一个创新的两阶段框架，称为CRVA-TGRAG（通过教师引导的检索-增强生成的脆弱性分析中的冲突解决）。首先，为了提高检索阶段的文档检索精度，我们采用了父文档分割和基于语义相似度和倒排索引的集成检索方案；其次，为了提高llm在生成阶段基于CVE数据集检索的能力，我们采用了教师引导的偏好优化技术对llm进行微调。我们的框架不仅通过RAG提高了内容检索的质量，而且还利用了llm中偏好微调的优势来更有效和准确地回答问题。实验表明，与外部知识库相比，该方法在检索最新cve方面具有更高的准确性和相关性。总之，我们的框架显著减轻了仅仅依赖法学硕士进行知识检索可能产生的潜在知识冲突和不一致。
Large Language Models (LLMs) are essential for analyzing and addressing vulnerabilities in cybersecurity. However, among over 200,000 vulnerabilities were discovered in the past decade, more than 30,000 have been changed or updated. This necessitates frequent updates to the training datasets and internal knowledge bases of LLMs to maintain knowledge consistency. In this paper, we focus on the problem of knowledge discrepancy and conflict within CVE (Common Vulnerabilities and Exposures) detection and analysis. This problem hinders LLMs' ability to retrieve the latest knowledge from original training datasets, leading to knowledge conflicts, fabrications of factually incorrect results, and generation hallucinations. To address this problem, we propose an innovative two-stage framework called \textbf{CRVA-TGRAG} (Conflict Resolution in Vulnerability Analysis via Teacher-Guided Retrieval-Augmented Generation). First, to improve document retrieval accuracy during the retrieval stage, we utilize Parent Document Segmentation and an ensemble retrieval scheme based on semantic similarity and inverted indexing. Second, to enhance LLMs' capabilities based on the retrieval of CVE dataset in generation stage, we employ a teacher-guided preference optimization technique to fine-tune LLMs. Our framework not only enhances the quality of content retrieval through RAG but also leverages the advantages of preference fine-tuning in LLMs to answer questions more effectively and precisely. Experiments demonstrate our method achieves higher accuracy in retrieving the latest CVEs compared to external knowledge bases. In conclusion, our framework significantly mitigates potential knowledge conflicts and inconsistencies that may arise from relying solely on LLMs for knowledge retrieval. 
\end{abstract}

\begin{figure}[t]
    \centering    
    \includegraphics[width=0.95\linewidth]{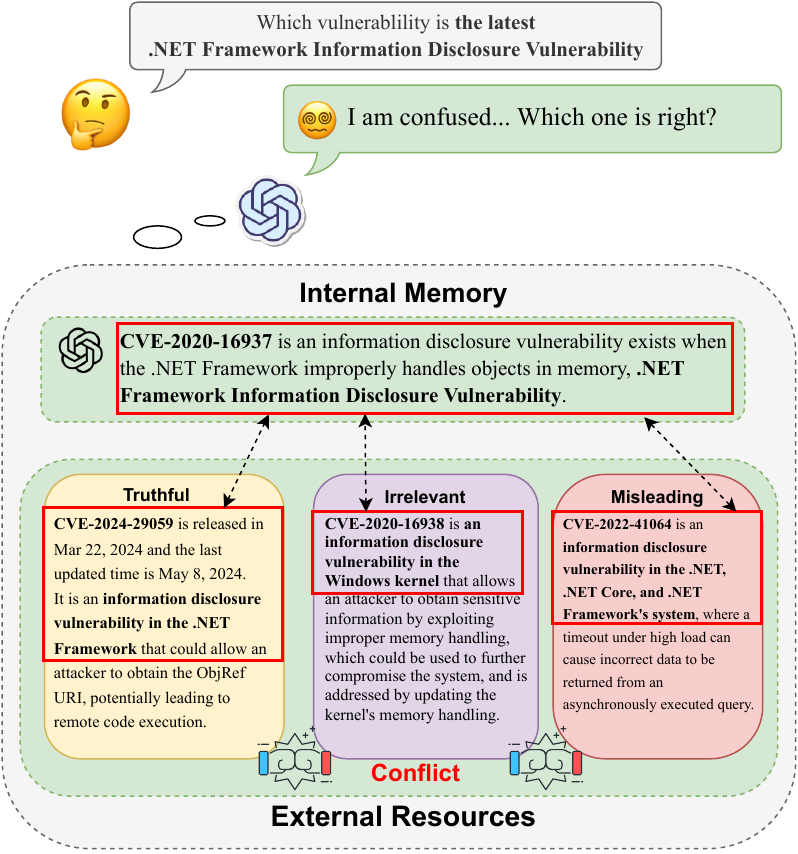}
    \caption{Knowledge conflict in LLM retrieving CVE related information}
    \label{fig:Knowledge Conflict}
\end{figure}

\section{Introduction}
% 以一个小故事开始
% 随着时间的流逝，世界上的事实和知识也在变化，甚至独自发展。例如，“美国当选总统”、“勒布朗·詹姆斯的主队”和“最近一届奥运会的主办城市” 等答案可能会在四年左右的时间里发生变化。此外，其他可能会更频繁和更快地进化，例如Python库（numpy， matplotlib和pandas等）和网络威胁情报（CTI）信息（CVE编号和CVSS文档）。这些知识的更新和升级通常分散在github存储库中的网站、项目文档甚至记录中。知识的这些特征对LLM提出了巨大的挑战，即如何跟随知识的演变和变化的路径，同步更新检索能力。
As time goes by, facts and knowledge about the world change and evolve alongside with them  \cite{roberts2020much,hu2023large,hu2024evaluating}. For instance, certain information, such as ``the President-elect of the United States", ``LeBron James's home team", and ``the latest Olympic host city", may change approximately every four years  \cite{jang2021towards}. Additionally, some information evolves more frequently and rapidly, such as Python libraries  \cite{code1} and Cyber Threat Intelligence (CTI) data  \cite{CTI1}, including CVE and CWE numbers. The updates and upgrades of this knowledge are often scattered across official websites like National Vulnerability Database (NVD)\footnote{https://nvd.nist.gov/}, project documentation, and even commit records in GitHub repositories, both explicitly and implicitly  \cite{xu2024knowledge}. These characteristics of knowledge present significant challenges for current LLMs in tracking the evolution and changes of knowledge to simultaneously enhance their retrieval capabilities  \cite{chen2023say}.

% 现有的方法有哪些
% 与传统机器学习模型相比，大语言模型经过超大规模文本数据的预训练，能够学习到更为丰富的通用知识。以OpenAI的ChatGPT为代表的闭源大语言模型，在通用任务求解、特定任务推理、指令遵循和与人类能力对齐方面表现日益出色。以LLAMA和Mistral为代表的开源大语言模型，进一步使用监督微调和偏好调优后，能够有效学习特定领域的知识，形成问答偏好，从而增强其生成和表达能力，显著提升解决特定领域复杂任务，回答复杂问题的能力。以检索查询CVE编号对应的漏洞信息为例，将CVE漏洞信息作为外部知识与数据，结合LLM的上下文进行输入，可以显著提高LLM在解答与CVE相关问题时的准确性。因此，在必要时将外部最新、准确的知识作为上下文提示信息输入LLM，能够有效缓解预训练阶段和知识更新过程中可能出现的知识冲突与不一致问题。
Compared to traditional machine learning models  \cite{hinton2006reducing}, LLMs  \cite{llm1} can learn extensive common sense knowledge after being pre-trained on large amount of datasets. Closed-source LLMs, such as GPTs  \cite{gpt-4}, have demonstrated impressive performance in solving general tasks  \cite{general-task}, following input instructions, aligning with human capabilities and reasoning for specific tasks. Open-source LLMs, such as Llama \cite{llama2} and Mistral  \cite{mistral}, can effectively learn domain-specific knowledge and develop question-answering preferences through supervised fine-tuning and preference fine-tuning. This significantly expands the generative and expressive capabilities of LLMs, enhancing their ability to generate responses and tackle various complex tasks. For instance, integrating CVE vulnerability data as external knowledge within the context of the LLM can significantly enhance the accuracy of the LLM's responses to CVE-related inquiries when people query vulnerability information associated with CVE-IDs. Therefore, associate the most accurate and the latest external knowledge as contextual prompt into LLMs could effectively alleviate potential knowledge conflicts and inconsistencies arising from pre-training phase and knowledge update process.

% 现有的方法有哪些问题：需要突出知识冲突以及冲突前后的比较问题
% CVE数据库的更新时间以小时为单位。受限于大语言模型知识库和训练数据集的时效性和大语言模型训练过程的较大代价，一个已经训练完成的大语言模型，在离线状态下暂时无法随时学习到当前CVE的最新知识。现有大语言模型通常存在“训练知识截止日期”，即大语言模型收集训练数据并进行训练的最晚时间。因此向大语言模型提出晚于该训练截止日期的问题，回答结果不一定正确，甚至会编造与事实不相符的答案，出现幻觉现象。模型微调技术的发展，使得LLM可以在特定领域或任务上对预训练模型进行进一步训练，使其在特定任务上的表现更加优化，却无法从根本上解决知识冲突问题。现有的传统RAG框架尽管能够联系外部知识库，对提出的问题在知识库中寻找到最佳匹配的答案后，作为上下文输入给LLM，然而知识库中可能存在自相矛盾的知识，在输入给LLM的上下文中可能隐藏有冲突，进而让LLM无法正确回答问题。同时，RAG框架对于文档的Split和embedding技术和方案选择不同，导致检索的质量参差不齐，最终影响回答问题的效果。Fayyazi 等人提出的 ProveRAG，借助RAG对现有的CVE内容进行检索增强，并对生成内容以匹配出处，证明该内容的准确性，能够一定程度上提高检索质量，然而该方案突出的是证明检索和生成内容的准确性，未能够很好比较CVE知识变动前后的区别。Fang等人使用LLM智能体自动挖掘零日漏洞的方案，侧重于研究LLM解决具体漏洞的能力，忽视了漏洞更新和漏洞变动对LLM产生的影响。
The update frequency of the CVE database is measured in hours. Due to the timeliness of the knowledge base and training datasets of LLMs, as well as the significant costs associated with the training process, an offline pre-trained LLM cannot learn the latest CVE knowledge in real-time. Existing LLMs typically have a ``knowledge cutoff date"  \cite{cutting_off} which refers to the latest time that the model finished training. Therefore, questions to the LLM after the date may yield incorrect answers, or even fabricate responses that are inconsistent with the facts, resulting in hallucinations. Figure \ref{fig:Knowledge Conflict} shows the conflict phenomenon in using LLM to retrieve knowledge related to ``The latest CVE Vulnerability". The advancement of fine-tuning techniques allows LLMs to further training on pre-trained models for specific domains or tasks, optimizing their performance in those specific Q\&A \cite{QA}. However, this method does not fundamentally resolve the issue of knowledge conflicts. Traditional Retrieval-Augmented Generation (RAG) method could retrieve external knowledge bases and match the best answers for the questions, which are then prompt as context to the LLMs. There may be contradictory knowledge within the knowledge base. This can lead to conflicts hidden in the context provided to the LLM, preventing it from answering questions accurately.

% 我们是怎么解决的
% 受到CVE知识和数据库频繁更新的限制，LLM无法通过频繁的训练和微调准确获取最新的知识。为了解决LLM在CVE领域的知识冲突问题，我们提出了一个二阶段的解决知识冲突的框架CRVA-TGRAG (Conflict Resolution in Vulnerability Analysis via Teacher-Guided Retrieval-Augmented Generation). 首先，我们注意到CVE数据来源之间存在冲突。CVE数据来源广泛，但是不少数据权威性不够，尚未与官方同步更新，存在不一致和冲突现象；其次是使用LLM进行CVE编号检索时，检索精度较低。相邻CVE编号内容极有可能相似性高，LLM无法准确辨别相邻相似CVE编号，造成LLM输出有误。面对以上两个问题，首先我们使用CVE官方提供的API和官方Github数据集，以Json格式下载CVE描述文档，整理出每一个CVE编号更新前和更新后的知识以及其区别。这样确保CVE内容的准确信和权威性，从数据集来源方面消除知识冲突的影响。其次，我们考虑使用RAG方案，通过检索上一阶段收集整理的外部数据集，配合LLM能够更准确生成内容。因此我们改进了RAG方案，针对于CVE文档的特性，我们对于文档的切分和嵌入过程进行优化，同时在查询阶段合并了多种查询方式以便于提高检索质量，更好检索到所需要的文档。我们希望LLM能够偏好于输出更新后的知识，因此我们将数据集中更新前和更新后的知识作为偏好调优的数据集，让LLM学习并产生偏好，接受更新后的知识，进一步解决知识冲突。

To address this problem, in this paper, we propose a two-stage framework for conflict resolution, named CRVA-TGRAG. In the course of our research, the first question we have faced is there are conflicts among the sources of CVE data. The sources of CVE data are extensive, but many lack sufficient authority and have not been updated with official sources synchronously, leading to inconsistencies and conflicts. Secondly, the retrieval accuracy is low when using LLM for CVE number searches. Adjacent CVE numbers are likely to have high similarity in content, making it difficult for LLM to accurately distinguish between them and resulting in incorrect outputs. Hence, we first use the API provided by NVD and CVE official GitHub dataset to download CVE documents in Json format, organizing the knowledge before and after updates for each CVE number and labeling their differences. It ensures the accuracy and authority of CVE content, eliminating the impact of knowledge conflicts from the dataset. Next, we consider using RAG approach. It could retrieve the external CVE dataset that collected in the previous phase and work with LLM to generate content more accurately. We improve the RAG approach, optimize the document segmentation and embedding processes according to the characteristics of CVE documents, while merging various query methods during the query phase to enhance retrieval quality and better access the required documents. Also, we hope LLM would prefer to generate updated knowledge. So we use the CVE knowledge before and after updates in the dataset as a preference-tuning dataset, allowing LLM to learn and develop preferences to accept updated knowledge, further resolving knowledge conflicts.

% \nese{First}, we utilize the official API provided by CVE to download the description documents for each CVE-ID, saving them in different file formats and selected the best format. We refer to Chiranjeevi's approach \footnote{https://github.com/morpheuslord/CVE-llm\_dataset} and organize the CVE documents that have undergone changes in time order. We further standardize the format to generate fine-tuning datasets and preference datasets for LLM fine-tuning. Next, we draw on the design principles of traditional RAG to differentiate document formats and optimize the document splitting and embedding processes. During the query phase, we combine multiple query methods to enhance retrieval quality. 
% 我们解决的还不错
% 我们对比前人的研究内容，设计评价指标评估了CRVA-TGRAG的生成质量，可靠性等指标。实验结果表明，我们的框架能够有效解决CVE更新对于LLM带来的影响，缓解了LLM在回答CVE相关问题时候产生的知识不一致和知识冲突问题，有效提高了LLM的生成质量，更有效帮助研究人员分析漏洞。
We compared our work with previous research and designed evaluation metrics to assess the generation quality and reliability of CRVA-TGRAG. The experimental results indicate that our framework effectively addresses the impact of CVE updates on LLMs, alleviating the issues of knowledge inconsistency and conflicts that arise when answering CVE-related questions. This, in turn, significantly enhances the generation quality of the LLM, providing more effective support for researchers in analyzing vulnerabilities. 

% 我们的贡献
% 我们的贡献有如下三点：
% （1）我们提出了一个以CVE数据为基础的偏好调优数据集和生成该数据集的方式
% （2）我们提出了一个二阶段的解决知识冲突的框架CRVA-TGRAG，有效缓解了LLM在面对CVE问题时候产生的知识不一致和知识冲突问题。
% （3）我们设计了多样化的评价指标，在消融实验中对基准方法和CRVA-TGRAG方法进行比较，评估结果表明我们的框架在生成内容的准确性，可靠性和忠实性上高于基准方法，证明了我们框架的有效性。
Our contributions are summarized as follows:
\begin{itemize}
\item We first address the gap in the field of vulnerability analysis for the lack of a knowledge conflict dataset. The dataset includes 1,260 pairwise conflict CVE items that have been published and modified over the past decade. It is accessible in our GitHub repository, accompanied by detailed explanations of the processing steps.

\item We present a two-stage framework named CRVA-TGRAG for knowledge conflict resolution. Our method enhances answer correctness by 27\% (increasing from 0.49 to 0.76) and faithfulness by 52\% (rising from 0.36 to 0.88) compared to the baseline. This framework effectively mitigates problems related to knowledge inconsistency and conflict, thereby validating our method.

% 传统的自然语言处理评价指标只能衡量摘要和机器翻译效果，对于评价漏洞检索方面的知识冲突存在局限性，因此我们使用LLM辅助评估的基础上，使用多维度评价指标评价LLM解决知识冲突的效果，将单一的指标针对框架或者方法性能的一个特定方面。这确保了指标既可解释又可操作，提供了对所测量内容的清晰见解。

\item We establish a scoring criteria that clearly describes the effectiveness of LLMs in resolving knowledge conflicts. We employ multidimensional metrics that consider various attributes, providing a detailed description of the LLM's  ability to resolve knowledge conflicts in vulnerability analysis. This method ensures that the metrics are both intuitive and interpretable, providing clear insights into measurement.

\end{itemize}

\section{Background \& Related work} 
\subsection{Knowledge Conflict}
Knowledge conflict  \cite{xu2024knowledge} refers to the discrepancies, contradictions, and conflicts between the internal parameterized knowledge of LLMs  \cite{xie2023adaptive} and externally acquired knowledge from contextual environments  \cite{tug-of-war}. This phenomenon affects the model's understanding and judgment of questions and can lead to outputs containing factual inaccuracies, logical inconsistencies, and other significant flaws, thereby diminishing the model's reliability and credibility  \cite{Time-aware}. The internal parameterized knowledge of LLMs is characterized by a clear temporal limitation. Once the model training is completed, this knowledge cannot be updated. As time progresses, the model's internal knowledge gradually becomes outdated and ineffective  \cite{cutting_off}. To compensate for this lag in internal knowledge, users often need to provide dynamic external supplementary information in practical applications, such as contextual dialogue history or relevant document retrieval  \cite{Resolving_knowledge}. However, contradictions and conflicts are often unavoidable when the model attempts to integrate external dynamic knowledge with its own outdated static knowledge, resulting in knowledge conflict.

\subsection{Retrieval Augmented Generation (RAG)}
% 检索增强生成（RAG）是一种将信息检索技术与llm相结合的新技术。它通过从外部知识库检索相关信息并使用提示来增强LLM处理知识密集型任务的能力。用户可以与LLM进行正常的会话交换，同时动态添加外部数据集进行检索，从而增强模型在特定领域的知识问答能力。Piktus等人最早提出了一个通用的基于微调的RAG方案，应用于知识密集型的NLP任务中，在开放领域的QA问答中取得了不俗成绩。人们在此基础上提出了上下文检索-增强生成（RAG），这是一种先进的RAG技术，它通过在检索过程中合并上下文压缩来提高响应的相关性和效率。在Contextual RAG中，检索到的文档在传递给语言模型之前通过Document Compressor进行处理。这个压缩器只提取和保留查询最相关的信息，甚至丢弃整个不相关的文档。这种方法减少了检索上下文中的噪声，从而从生成模型得到更精确、更简洁、更经济的响应。研究人员将RAG的检索阶段不断优化，提出了hybrid RAG 和 RAG Fusion 方法，进一步提高RAG检索正确率和效率。然而上述方法面对本文研究的CVE知识冲突问题解决效果不够理想，检索的信息准确率没有达到预期效果，仍然有值得优化和改进的空间。
RAG is a novel technology that combines information retrieval techniques with LLMs. It enhances LLMs' ability to handle knowledge-intensive tasks by retrieving relevant information from external knowledge bases and using as prompts, enhancing the model's knowledge question-answering capabilities in specific domains. Researchers firstly propose a general fine-tuning-based RAG approach, which was applied to knowledge-intensive NLP tasks and achieved notable results in open-domain question answering  \cite{RAG}. Building on this foundation, the concept of Contextual-RAG\footnote{https://www.anthropic.com/news/contextual-retrieval} was introduced, representing an advanced RAG technique that improves the relevance and efficiency of responses by merging context compression during the retrieval process. Continuously, researchers optimized the splitting, embedding retrieval stage of RAG, proposed Hybrid RAG  \cite{hybrid_rag1_yuan} and RAG Fusion  \cite{rag_fusion} to further improve the accuracy and efficiency. However, regarding the CVE knowledge conflict problem mentioned earlier, the aforementioned RAG methods are not ideal and the accuracy of the retrieved information does not meet expectations.

\subsection{LLMs in Vulnerability Retrieval and Assessment} 
Existing work has shown that LLM has made good progress in software vulnerability detection. At the same time, LLM is also used for the evaluation and analysis of searches. Dunlap et.al  \cite{dunlap2024pairing} addresses the increasing complexity of managing security vulnerabilities within open-source software dependencies. Hu el.al proposed DeGPT  \cite{hu2024degpt}, which is an end-to-end framework for optimizing decompiler output based on LLMs. And Fayyazi et al. introduced Provenance RAG  \cite{ProveRAG} to validate the retrieval accuracy in proposing provenance from website and official knowledge base. Most of the above methods are to detect, analyze and evaluate the specific program vulnerability code represented by CVE, and do not involve the vulnerability knowledge of CVE itself and the study of knowledge conflict. Therefore, our paper will focus on LLM to resolve the conflict knowledge in CVE vulnerability analysis.

\section{Motivation \& Research Questions} %\zzy{Finished}
%Motivation这里要改一下，应该是我们受到什么启发然后来做这个framework 强调冲突和时效性的问题 Highlevel idea RQ下面的文字解释RQ并略写 突出RQ的重要性
% 解决大语言模型中的知识冲突问题，传统方法通常通过使用特定领域的静态数据集对大语言模型进行监督式微调，从而提升其在特定领域知识问答中的准确性，减少知识冲突。此外，也有研究引入了检索增强生成（RAG）方案，通过结合外部数据集检索，将其作为上下文输入到大语言模型中。然而，微调方法的计算量和系统开销较大，尤其在面对频繁更新的CVE知识时，微调方案的成本显得尤为高昂。传统的RAG方案在处理大量相似的CVE编号和漏洞描述时，检索的准确率较低，容易出现错误。同时，现有的多种漏洞分析方案虽然能够评估漏洞，但往往无法提供有关漏洞更新前后变化的有效见解和分析。漏洞的快速更新引发了大语言模型的知识冲突，这对大语言模型的应用构成了挑战。基于这些挑战，本研究的动机便来源于此，我们旨在通过回答以下研究问题来解决这些问题：
To address knowledge conflicts in LLMs, traditional methods typically adapt supervised fine-tuning and use static datasets specific to certain domains. This approach aims to enhance the accuracy of knowledge question-answering in those domains while reducing knowledge conflicts. Additionally, some studies have introduced RAG that combines external dataset retrieval and use it as contextual input for LLMs. However, the computational demands and system overhead associated with fine-tuning methods can be significant, particularly when dealing with frequently updated CVE knowledge, resulting in high fine-tuning costs. Traditional Naive RAG approaches often exhibit low retrieval accuracy when processing numerous similar CVE identifiers and vulnerability descriptions, leading to potential errors. Furthermore, while existing vulnerability analysis solutions can assess the presence of vulnerabilities, they often fail to provide effective insights and analyses regarding changes before and after vulnerability updates. The rapid pace of vulnerability updates has exacerbated knowledge conflicts in LLMs, posing significant challenges for their application. Motivated by these challenges, our research aims to address the following questions:

\begin{figure*}[h]
    \centering
    \includegraphics[width=1\linewidth]{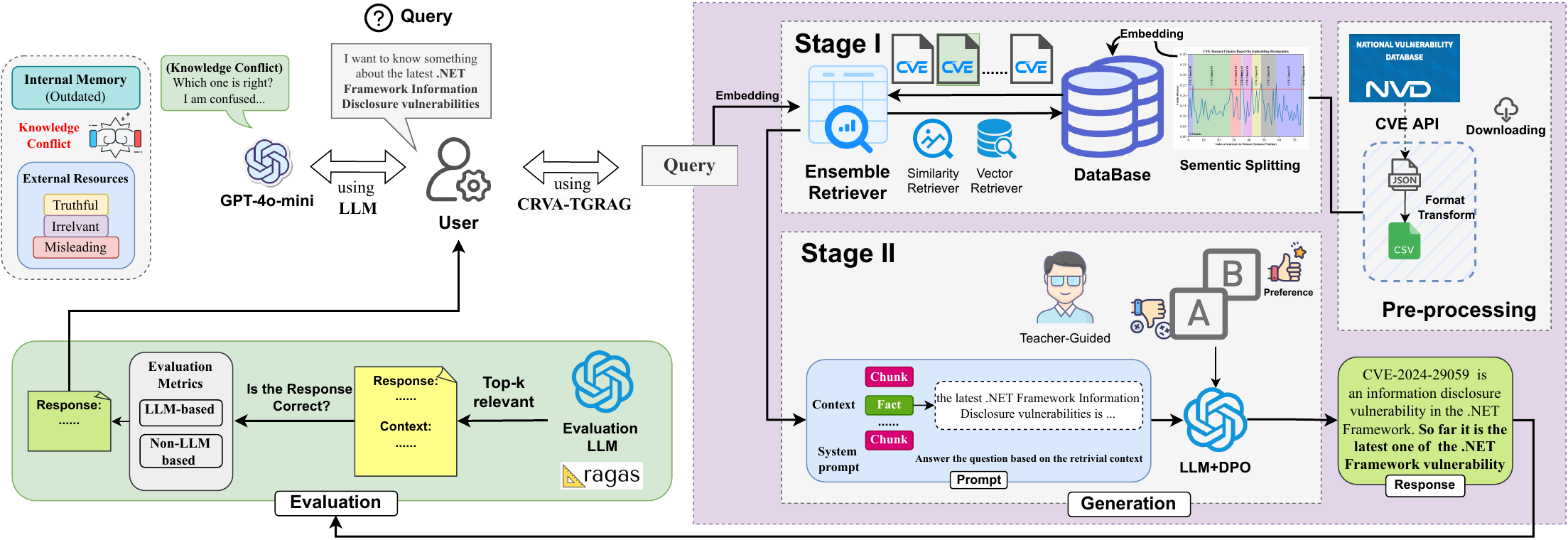}
    \caption{The pipeline of our framework. The user queries GPT-4o-mini, but due to internal knowledge conflicts, the returned results are incorrect. In our method, we obtain the latest CVE knowledge from official sources during pre-processing. In \textbf{Stage I}, we segment the formatted files using a semantic similarity-based segmentation approach and perform retrieval using ensemble retrieval. Then in \textbf{Stage II}, we fine-tune the LLM through teacher-guided preference optimization. Meanwhile, we evaluate the output's accuracy, faithfulness, and other metrics through our evaluation framework.}
    \label{fig:pipeline}
\end{figure*}

% 如何解决LLM内部未更新的知识与外部更新数据不一致所导致的CVE知识冲突问题，并区别出更新之后的区别呢？
% 受到模型训练截止时间的限制，更新后的CVE信息与LLM原有的训练数据集中的CVE信息产生不一致，在使用LLM查询变动的CVE信息时候发生知识冲突。在冲突的同时LLM无法获取变动CVE信息的内容，生成不一致的内容，甚至出现幻觉现象，在实际应用中，这种现象干扰误导用户。因此数据集来源需要从官方权威的数据集中获取，这样才能确保数据的时效性和准确性。

% 为了解决这个问题，我们使用官方同步更新的CVE数据库作为数据集来源，确保数据正确性和准确性。同时，我们借助NVD的API获取到产生变更的CVE信息，确保获取信息渠道的权威性，最后我们使用llm将变更前和变更后的CVE信息进行比较。生成对应的Json文档，格式为键值对形式{"CVE-ID", "OldValue", "NewValue", "Changed details"}.
\textbf{RQ1: How to solve the problem of CVE knowledge conflict caused by the inconsistency between LLM internal unupdated knowledge and external updated data, and distinguish the difference after updating?}
Due to LLM's inherent ``knowledge cutoff date" the updated CVE information is inconsistent with the old one in the original training dataset, leading to knowledge conflicts when LLM is queried details of the changed CVE information. Simultaneously, LLM cannot obtain the contents of the changing information, generate inconsistent content, and may even hallucinate in response. Practically, this problem would disturb adn misguide users. Therefore, the dataset must be sourced from official and authoritative datasets to ensure the timeliness and accuracy of the data. 

% RQ2: CVE编号数字接近，不少相邻的CVE编号漏洞内容相似，如何设计一个能检索CVE漏洞信息的RAG系统，提高检索精度和检索质量呢？
% RAG系统包括四个主要部分：indexing, retriever, Augment and Generation. 提高各部分的性能可以使得整体RAG的检索质量得到提升，例如修改分块方式，改进检索方法等。因此我们的框架改进了RAG方案，提高检索质量，这对于系统整体性能的提升起到较大作用和积极影响。
%% 在Index阶段我们改进了文档分割模式，大量CVE文档信息进行分割时候，同一个CVE文档内容需要被分割在同一文本块中进行embedding。在检索阶段，需要基于用户提出的问题检索到最相关的文档，受到RAG的启发，我们改进了检索方式，采用语义相似度与向量检索结合的方式，同时将输入的查询进行同义替换，生成子问题的联合查询，使得检索器能更准确检索到与问题相关程度高的文档，提高检索准确性
\textbf{RQ2: How can we design a RAG system to enhance retrieval accuracy and quality?} Traditional RAG system consists of four main components: Indexing, Retriever, Augmentation and Generation. Enhancing the performance of each component could improve the overall retrieval quality of the system, such as modifying the splitting mode and refining the retrieval method. CVE numbers are often similar, and many adjacent CVE numbers contain related information. The retrieval performance of traditional RAG schemes performs inaccurate when faces with a large number of similar CVE documents. Therefore, it is necessary to design an optimized RAG scheme based on the characteristics of CVE documents to enhance retrieval accuracy.

% RQ3：在检索内容作为最终输入上下文的基础上，如何进一步提高输出质量，改进输出内容，更好解决由于检索内容产生的冲突问题
% 一般情况下，生成阶段使用标准聊天类大语言模型，接受检索到的数据作为上下文，结合上下文和查询内容进行输出。我们改进了使用的大语言模型，将CVE新知识作为偏好内容对大语言模型进行微调，使其产生针对特定问题的问答偏好并倾向于回答偏好内容，提高了生成的准确性，缓解知识冲突

\textbf{RQ3: Based on the retrieved content as the input context, how to further improve the quality of the output content and better address the knowledge conflicts caused by the retrieved content?} 
RAG uses a standard chat-LLM during the generation stage and accepts the retrieved data as context. The output is combined with both the context and the query content. We enhance the chat-LLM in generation stage and fine-tune it by incorporating the updated CVE knowledge as the preferred content, which also improves generation accuracy and alleviates knowledge conflicts.

\section{Methodology}
% 这里加入两个intuition 看一下23年CCS那篇文章怎么写的 先写Intuition，再写methodology
\subsection{Key Intuition}
% 受到学生阅读教材的偏好和教师指导学生学习教材的启发，我们针对RQ1-RQ3提出了下面的直觉：
% Intuition 1: 未经过排版的教材与格式化排版整齐的教材相比，学生更倾向于阅读格式化排版整齐的教材。
% intuition 2: 教材各章节结构层次清晰，重点突出，更有利于学生学习掌握
% Intuition 3: 新版本教材出版后，教师直接指导学生比较新旧教材内容的差异，着重学习教材内容变更的部分，而不是单独让学生自学新教材内容，更有利于学生较快掌握知识点

Inspired by students' preferences for reading materials from textbook with teachers' guidance, we propose the following intuitions regarding \textbf{RQ1-RQ3}:

\begin{itemize}
    \item \textbf{Intuition 1:} Students are more inclined to read well-formatted materials to grasp knowledge compared to unformatted ones.
    \item \textbf{Intuition 2:} A clear structure and emphasis on key points in each chapter of the material facilitate students better learning and understanding.
    \item \textbf{Intuition 3:} After the publication of a new version of the textbook, teachers should guide students to compare the differences between the old and new content, focusing on the changes rather than letting them study independently, to help them grasp knowledge more quickly and accurately.
    \end{itemize}
\subsection{CRVA-TGRAG Framework}

% 基于上述直觉，我们将学生学习知识的过程与LLM学习外部知识的过程进行类比，进而引出设计我们方法设计。我们结合上述直觉和分析，将CRVA-TGRAG Framework设计为两个阶段的框架，下述将分别介绍两个阶段的的具体方法：
Based on the intuitions, we draw an analogy between the process of students learning knowledge and the process of LLMs acquiring external knowledge, leading to design our method. We combine the above intuitions and analyses to design the CRVA-TGRAG as a two-stage framework. Figure \ref{fig:pipeline} shows the pipeline of CRVA-TGRAG The following subsections will introduce the specific methods for each stage.

\subsubsection{Data Pre-processing (Preliminary)}\label{subsubsec:Preliminary}
%基于直觉1，为了解决RQ1的问题，我们使用官方同步更新的CVE数据库作为数据集来源，这样确保数据正确性和准确性。同时，我们借助NVD的API获取到产生变更的CVE信息，确保获取信息渠道的权威性。我们将下载的CVE信息由Json文档转换为为CSV文档，每一个CVE信息对应一个CSV文档。格式化的CSV文档相比于Json文档而言，添加元数据更方便，这样为后面的文档分块嵌入和检索提供便利。最后我们使用llm将变更前和变更后的CVE信息进行比较，生成对应的Json文档，格式为键值对形式，如下表所示
To solve \textbf{RQ1}, based on \textbf{Intuition 1} we use the officially synchronized CVE database\footnote{https://github.com/CVEProject/cvelistV5} as the source of our dataset, ensuring the correctness and accuracy of the data. At the same time, we utilize the NVD API\footnote{https://nvd.nist.gov/developers/vulnerabilities} to obtain the CVE information that has undergone changes, ensuring the authority of the information source. We convert the downloaded CVE information from Json documents to CSV documents, with each CVE information corresponding to a separate CSV document. Formatted CSV documents are more convenient for adding metadata compared to Json documents, facilitating chunk splitting, embedding and retrieval subsequently. Finally, we design specific prompts to guide LLM compare the CVE information before and after the changes, generating the corresponding Json document. The following is the prompt and the format of the Json document:

\begin{mybox}[colback=yellow!6!white]{Generation Prompt}
{ \small \it 
    You are Bob, a cybersecurity professor and expert. Please compare the differences in the specific content of the same CVE number after changes and summarize the changes in several sentences. Remain faithful to the original text and do not fabricate content.
}
\end{mybox}

\begin{lstlisting}[language=Json,basicstyle=\fontsize{6}{8}\ttfamily]
{
    "CVE-ID":
    "OldValue":
    "NewValue":
    "Changed details":
}
\end{lstlisting}

\subsubsection{Retrieval and Augmentation (Stage I)}
% 通常RAG系统包括四个部分：indexing, retriever, Augment and Generation。提高各部分的性能可以使得整体RAG的检索质量得到提升。在Indexing阶段需要将所有数据预处理阶段的CSV文档信息通过文档加载器加载合并后统一进行分割，同一个CVE文档内容最好被分割在同一文本块中进行embedding，因此我们需要改进文档分割模式。基于直觉2的启发，我们希望在长文档的分块中，每一块中拥有足够长的文档，以便保留每个块完整的上下文，同时我们希望能够同时使用较小的文档，以便它们的嵌入可以更准确。同时较小文档仍然有可能包含了相邻两个CVE-ID的信息，在此基础上我们使用基于语义相关的文档分块形式，确保同一个CVE-ID的信息能够尽量包含在同一块中。最终在分块后将CVE-ID作为元数据添加到父文档中，便于之后的检索工作。
% 在检索阶段，系统需要基于用户提出的问题检索到最相关的一个或者几个文档，受到RAG的启发，我们改进了检索方式，采用BM25语义相似度与向量检索组合的方式进行检索。我们检索的问题中直接包含有CVE-ID，因此我们将组合检索器的检索权重设置为[0.25,0.75]，我们将输入的查询进行同义替换，生成子问题的联合查询，使得检索器能更准确检索到与问题相关程度高的文档，提高检索准确性。
Typically, a RAG system consists of four components: indexing, retrieving, augmentation, and generation. Improving the performance of each component can enhance the overall retrieval quality of the system. During the indexing phase, it is required to load and merge all CSV documents from \textbf{\nameref{subsubsec:Preliminary}} using a document loader, and then uniformly segment it. Ideally, the content of the same CVE document should be segmented into the same text chunk for embedding, which necessitates improvements in the document segmentation model. Inspired by \textbf{Intuition 2}, we aim to ensure that each chunk of a long document contains sufficient length to retain complete context, while also allowing for smaller documents to enable more accurate embeddings. Additionally, smaller documents may still contain information from adjacent CVE-IDs. Based on this, we employ a semantically relevant document chunking approach to ensure that information related to the same CVE-ID is included in the same chunk whenever possible. Finally, after chunking, we add the CVE-ID as metadata to the parent document to facilitate subsequent retrieval tasks.

During the retrieval phase, the system needs to retrieve the most relevant one or several documents based on the user's questions. Inspired by RAG \cite{rag_fusion}, we improved the retrieval method by combining BM25  \cite{BM25} semantic similarity with vector retrieval. Our retrieval questions directly include CVE-IDs, so we set the retrieval weights of the combined retriever [vector,semantic] to [0.25, 0.75]. We perform synonym replacement on the input queries to generate joint queries for sub-questions, enabling the retriever to more accurately retrieve documents that are highly relevant to the questions, thus improving retrieval accuracy.
\begin{equation} \label{formula:BM25}
    \resizebox{0.9\linewidth}{!}{$
    \displaystyle
    Score_{BM25}(D,Q)= \sum_{q_i \in Q}IDF(q_i) \cdot \frac{f(q_i,D) \cdot (k_1+1)}{f(q_i,D) + k_1 \cdot (1-b+b\cdot \frac{|D|}{avgDL})}$}
\end{equation}

\begin{equation}\label{formula:IDF}
    \resizebox{0.6\linewidth}{!}{$
    \displaystyle
    IDF(q_i)=\log{\left(\frac{N-n(q_i)+0.5}{n(q_i)+0.5}+1.0 \right)}
    $}
\end{equation}
where in formula (\ref{formula:BM25}),$f(q_i,D)$ is the frequency of query term $q_i$ in document $D$. $|D|$ is the document length. $avgDL$ is the average document length across the entire document dataset. $k_1$ and $b$: are tuning parameters, where $k_1$ is typically between 1.2 and 2, and $b$ is usually set to 0.75. And $IDF(q_i)$ is the inverse document frequency of the query term $(q_i)$ calculated as formula (\ref{formula:IDF}). In formula (\ref{formula:IDF}), where $N$ is the total number of documents, and $n(q_i)$ is the number of documents containing term $q_i$.

\subsubsection{Generation (Stage II)}
% 生成阶段通常使用聊天LLM，这一类LLM能够更好遵循指令并结合上下文进行回答。基于intuition3, 为了解决RQ3提出的问题，我们对聊天LLM进行进一步处理。Data Pre-processing阶段准备好的变更前和变更后的CVE信息是经过清洗质量较好的数据集，其中的Newvalue键对应的CVE内容是我们更需要的，将Json文档调整格式形成偏好调优数据集，使用DPO对LLM进行偏好调优，让LLM能够更偏好于更新后的CVE数据，有效减少了LLM对于CVE旧知识的偏好程度。
The generation stage typically utilizes chat LLMs, which are better at following instructions and responding in context. Based on \textbf{Intuition 3}, to address the issues raised in \textbf{RQ3}, we further process the chat LLM. The CVE information before and after changes, prepared during the data pre-processing phase, is a cleaned and high-quality dataset. The CVE content corresponding to the \textit{New Value} key is what we need most. We adjust the format of the Json documents to create a preference-tuning dataset and use DPO  \cite{DPO} to fine-tune the LLM. This allows the LLM to prefer the updated CVE data more, effectively reducing its preference for outdated CVE knowledge. The process of DPO is \textbf{Algorithm \ref{alg:DPO}}
% 我们的DPO算法如下所示

\begin{algorithm}[htbp]
% \caption{Offline Preference Optimization for Language Model}
\caption{Stage II DPO}
\begin{algorithmic}[1] \label{alg:DPO}
\STATE \textbf{Input:} A prompt \( x \), reference policy \( \pi_{\text{ref}} \), offline preference dataset \( D = \{(x^{(i)}, y_w^{(i)}, y_l^{(i)})\}_{i=1}^{N} \), and a desired hyperparameter \( \beta \)
\STATE \textbf{Step 1:} For each prompt \( x \), sample \( y_1, y_2 \sim \pi_{\text{ref}}(\cdot \mid x) \)
\STATE \textbf{Step 2:} Annotate the preferences of the samples \( y_1 \) and \( y_2 \) using human preferences, then build the offline preference dataset:
\[
D = \{(x^{(i)}, y_w^{(i)}, y_l^{(i)})\}_{i=1}^{N}
\]
where \( y_w^{(i)} \) and \( y_l^{(i)} \) represent the winning and losing choices respectively for each \( x^{(i)} \)
\STATE \textbf{Step 3:} Given \( \pi_{\text{ref}} \), dataset \( D \), and the desired \( \beta \), optimize the language model \( \pi_{\theta} \) (the policy) to minimize the loss function \( \mathcal{L}_{\text{DPO}} \)
\end{algorithmic}
\end{algorithm}

% 我们的优化目标是在最大化偏好数据一致性的同时，保持与参考模型（通常是经过监督微调的模型）的相似性。目标函数为

\begin{equation}\label{formula:DPO optimized target}
\resizebox{0.9\linewidth}{!}{$
    \displaystyle
    \max_{\pi_\theta}\mathbb{E}_{x \sim D, y \sim \pi_\theta(y|x)}[r_\phi(x, y)] -
    \beta D_{\text{KL}}(\pi_\theta(y | x) \| \pi_{\text{ref}}(y | x)) $}
\end{equation}

\begin{equation}\label{formula:DPO Loss Function}
\resizebox{0.9\linewidth}{!}{$
    \displaystyle
    \mathcal{L}_{DPO}(\pi_\theta; \pi_{\text{ref}}) = -\mathbb{E}_{(x, y_w, y_l) \sim D}\left[\log\sigma\left(\beta \log\frac{\pi_\theta(y_w|x)}{\pi_{\text{ref}}(y_w|x)} - \beta \log\frac{\pi_\theta(y_l|x)}{\pi_{\text{ref}}(y_l|x)}\right)\right]
    $}
\end{equation}

\begin{equation}\label{formula:DPO Gradient Update}
\resizebox{0.9\linewidth}{!}{$
\displaystyle
\nabla_\theta \mathcal{L}_{DPO}(\pi_\theta; \pi_{\text{ref}}) = -\beta \mathbb{E}_{(x, y_w, y_l) \sim D}\left[\sigma(\hat{r}_\theta(x, y_l) - \hat{r}_\theta(x, y_w)) \left(\nabla_\theta \log \pi_\theta(y_w | x) - \nabla_\theta \log \pi_\theta(y_l | x)\right)\right]$}
\end{equation}

\subsection{Evaluation Metrics} \label{Section:Evaluation Metrics}
% 我们的系统引入了RAG方法的同时，对输出部分的LLM进行偏好微调，为了综合评价我们方案的性能，我们在传统的语义评价指标例如BLEU ROUGE-L上，增加了评价RAG的相关指标。我们参考了RAGAS \cite提供的相关评估指标进行评估。在本章节中我们将介绍常规的基于LLM的评价指标和与RAG相关的评价指标。
Our system introduces the RAG method while also performing preference-tuning on the chat-LLM in the generation stage. To comprehensively evaluate the performance of our framework, we added relevant metrics for evaluating RAG to traditional semantic evaluation metrics such as BLEU \cite{BLEU} and ROUGE-L  \cite{ROUGE}. We refer to the relevant evaluation metrics provided by RAGAS\footnote{https://github.com/explodinggradients/ragas} \footnote{https://docs.ragas.io/} for our work. In this section, we will introduce the conventional natual language evaluation metrics and the metrics related to RAG evaluation. The up arrow after each evaluation metric indicates that the higher the positive correlation score, the better. And the down arrow, vice versa.

\subsubsection{Natural Language Metrics}
% answer_correctness 对答案正确性的评估包括衡量生成的答案与真实情况相比的准确性。这种评估依赖于基本事实和答案，得分范围从0到1。得分越高，表示生成的答案与基本事实之间的一致性越强，这意味着正确程度越高。答案正确性包括两个关键方面：生成的答案与基础真理之间的语义相似性，以及事实相似性。
% answer_relevancy “答案相关性”侧重于评估生成的答案与给定提示的相关性。不完整或包含冗余信息的答案得分越低，得分越高表示相关性越好。这个度量是使用问题、上下文和答案来计算的。答案相关度定义为原始问题与许多人工问题的平均余弦相似度，如下式所示
% answer_similarity：答案语义相似度的概念是指对生成的答案与基本真理之间的语义相似度进行评估。这种评估是基于基本事实和答案，其值在0到1的范围内。得分越高，表示生成的答案与基本事实之间的一致性越好。测量答案之间的语义相似性可以为生成的响应的质量提供有价值的见解。
(1)\textbf{Answer Correctness (AC$\uparrow$)}: It measures how accurately the generated answer aligns with the ground truth and bases on fundamental facts, with scores ranging from $(0,1)$. A higher score indicates a stronger alignment with the facts, signifying greater correctness. 

(2) \textbf{Answer Relevance (AR$\uparrow$)}: It measures the relevance of the generated answer to the given prompt. Answers that are incomplete or contain redundant information receive lower scores, while higher scores indicate better relevance. The expression of AR is defined as the average cosine similarity between the original question and many manually crafted questions.

\begin{equation}\label{formula:AR}
    \resizebox{0.85\linewidth}{!}{$
    \displaystyle
    AR=  \frac{1}{N} \sum_{i=1}^{N} cos(E_{gi},E_o) = \frac{1}{N} \sum_{i=1}^{N} \frac{E_{gi} \cdot E_o}{||E_{gi}||\cdot ||E_o||}
    $}
\end{equation}
where $E_{gi}$ and $E_{o}$ are the embedding of the generated question and original question separately and $N$ is the number of generated questions.

(3) \textbf{Answer Similarity (AS$\uparrow$)}: It measures between the generated answer and the ground truth. and bases on facts and answers, with values ranging from $(0,1)$. A higher score indicates better consistency between the generated answer and the facts.

% faithfulness：这是根据给定的上下文来衡量生成的答案的事实一致性。它是根据答案和检索到的上下文计算的。答案被缩放到（0,1）范围。越高越好。如果在答案中提出的所有主张都可以从给定的上下文中推断出来，则认为生成的答案是可靠的。为了计算这个值，首先从生成的答案中识别一组声明。然后，将这些声明中的每一个与给定的上下文进行交叉检查，以确定是否可以从上下文推断出来。
% context precision上下文精度是一种度量，用于评估上下文中存在的所有与基本事实相关的项目是否排名更高。理想情况下，所有相关的部门都必须出现在高层。该度量是使用问题、ground_truth和上下文计算的，其值范围在0到1之间，分数越高表示精度越好
% context recall：上下文回忆衡量检索上下文与注释答案一致的程度，被视为基本事实。它是使用问题、基本事实和检索的上下文计算的，值的范围在0到1之间，值越高表示性能越好。为了从基本真理答案中估计上下文召回，分析基本真理答案中的每个主张，以确定它是否可归因于检索的上下文。在理想的情况下，基础真理答案中的所有主张都应归因于检索到的上下文。
\subsubsection{RAG Metrics} 

(1) \textbf{Faithfulness (Faith$\uparrow$)}: It measures the factual consistency of the generated answer based on the given context. The answer is scaled to $(0,1)$. The higher the score, the better. The generated answer is considered reliable if all claims made in the answer can be inferred from the given context.

\begin{equation}\label{formula:Faithfulness}
    \resizebox{0.28\linewidth}{!}{$
    \displaystyle
    % F_{score}= \frac{|NumClaimsInferred|}{|TotalClaims|}
    F_{score}= \frac{|NCI|}{|TC|}
    $}
\end{equation}
where $NCI$ means \textbf{\textit{Number of Claims}} in the generated answer that can be \textbf{\textit{Inferred}} from given context and $TC$ means \textbf{\textit{Total number of Claims}} in the generated answer.

(2) \textbf{Context Precision (CP$\uparrow$)}: It measures whether all the contexts are relevant to the facts are ranked higher or not. Ideally, all relevant items should appear at higher levels. The value ranging from $(0,1)$. A higher score indicates better precision.

\begin{equation}\label{formula:CP}
    \resizebox{0.55\linewidth}{!}{$
    \displaystyle
    CP@k = \frac{\sum_{k=1}^{K} (Precision@k\cdot{v_k})}{|TopK|}
    $}
\end{equation}

\begin{equation}\label{formula:Precision}
    \resizebox{0.55\linewidth}{!}{$
    \displaystyle
    Precision@k = \frac{TP@k}{TP@k+FP@k}
    $}
\end{equation}
where $TopK$ is the total number of relevant items in the top K results, $K$ is the total number of chunks in contexts and $v_k\in\{0,1\}$ is the relevance indicator at rank $k$.

(3) \textbf{Context Recall (CR$\uparrow$)}: It measures the degree to which the retrieved context is consistent with the annotated answer, regarded as the fact. The values ranging from $(0,1)$. A higher value indicates better performance. To estimate context recall from the ground truth answer, each claim in the ground truth answer is analyzed to determine if it can be attributed to the retrieved context. Ideally, all claims in the ground truth answer should be attributable to the retrieved context.

\begin{equation}\label{formula:CR}
    \resizebox{0.23\linewidth}{!}{$
    \displaystyle
    CR = \frac{|C_{GT}|}{|N_{GT}|}
    $}
\end{equation}
where $C_{GT}$ means ground-truth that can be attributed to context and $N_{GT}$ means the number of claims in ground-truth.

\section{Experiments}
% 结合intuition回答每个RQ，我们首先设计并构建了一个漏洞分析领域的高质量的数据集（RQ1），之后我们构建了CRVA-TGRAG框架，基于RAG框架并使用基于教师指导的偏好微调，以提升系统对于知识冲突的解决能力（RQ2，RQ3）。最后，我们整合设计评估指标，对系统解决知识冲突的能力和生成效果进行有效评估（RQ3），实验证明了我们的系统在各项指标评估中表现较好，可以有效解决本领域中知识冲突问题
Combining intuitions to address each RQ, we first designed and constructed a high-quality dataset in the field of vulnerability analysis (\textbf{RQ1}). Next, we built the CRVA-TGRAG framework, which is based on the RAG framework and utilizes teacher-guided preference fine-tuning, to enhance the system's ability to resolve knowledge conflicts (\textbf{RQ2}, \textbf{RQ3}). Finally, we integrated and designed evaluation metrics to effectively assess the system's ability to resolve knowledge conflicts and its generation performance. Experiments demonstrated that our system performs well across various evaluation metrics and can effectively address knowledge conflict issues in this field.

\subsection{Dataset}
% 为了解决RQ1的问题，我们受到描述CVE的内容的json文件启发，我们获取了2024年1月1日至2024年12月31日为止的所有CVE信息，以json文件形式保存。我们注意到在这个时间段内官方发布的CVE编号中，仍有不少编号标注“修改”字段，即这类编号修改后发布，这一类编号有的是2023年甚至更早时间段的CVE编号。我们对数据进行进一步筛选，选出CVE编号中“baseSeverity”=HIGH的CVE编号，即这些漏洞被标注为高危漏洞。选择此类漏洞作为数据集更有助于研究人员关注研究高危漏洞问题。截止2024年12月31日，我们所搜集到的CVE编号年份位于2024年之前的共计3332条CVE编号，其中“baseSeverity”=HIGH的编号一共1060个，因此我们最终选择这一部分编号作为数据集进行处理。数据集整理的结果如下图所示：

\begin{figure}[htbp]
    \centering
    \includegraphics[width=0.85\linewidth]{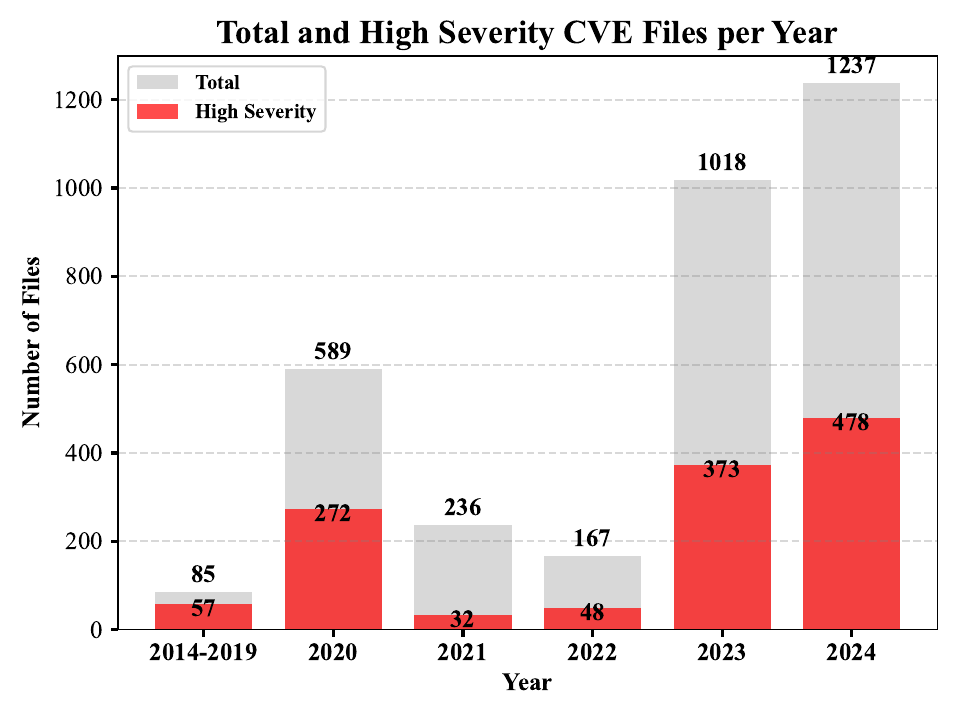}
    \caption{CVE vulnerabilities that have changed from 2014 to 2024. The gray parts represent the total number of changes that occurred in the corresponding year, The red parts indicate the number of high-risk CVE vulnerabilities that changed.}
    \label{fig:CVE_count}
\end{figure}

To address \textbf{RQ1}, we were inspired by the describe content in CVE Json Files. We collected all CVE files from Jan.1, 2024, to Dec.31, 2024. We noticed that among the CVE numbers officially released during this period, many were marked with a "modified" field, indicating that these numbers were released after modification, and some of these numbers date back to 2023 or even earlier. We further filtered the data to select CVE numbers where the key ``baseSeverity" == \textit{HIGH}, meaning these vulnerabilities are classified as high-risk vulnerabilities. Selecting such vulnerabilities as the dataset is more beneficial for researchers focusing on high-risk vulnerability issues. As of Dec.31, 2024, we collected a total of 3,332 CVE numbers from years prior to 2024, of which 1,060 numbers had ``baseSeverity" == \textit{HIGH}, Therefore, we ultimately selected these CVE for dataset processing. The results of the dataset organization are shown in the Figure \ref{fig:CVE_count} .

\subsection{Chunk Splitting}
\begin{figure*}[ht] 
\label{fig:chunk}
\centering
\begin{subfigure}[b]{0.32\textwidth}
    \includegraphics[width=\textwidth]{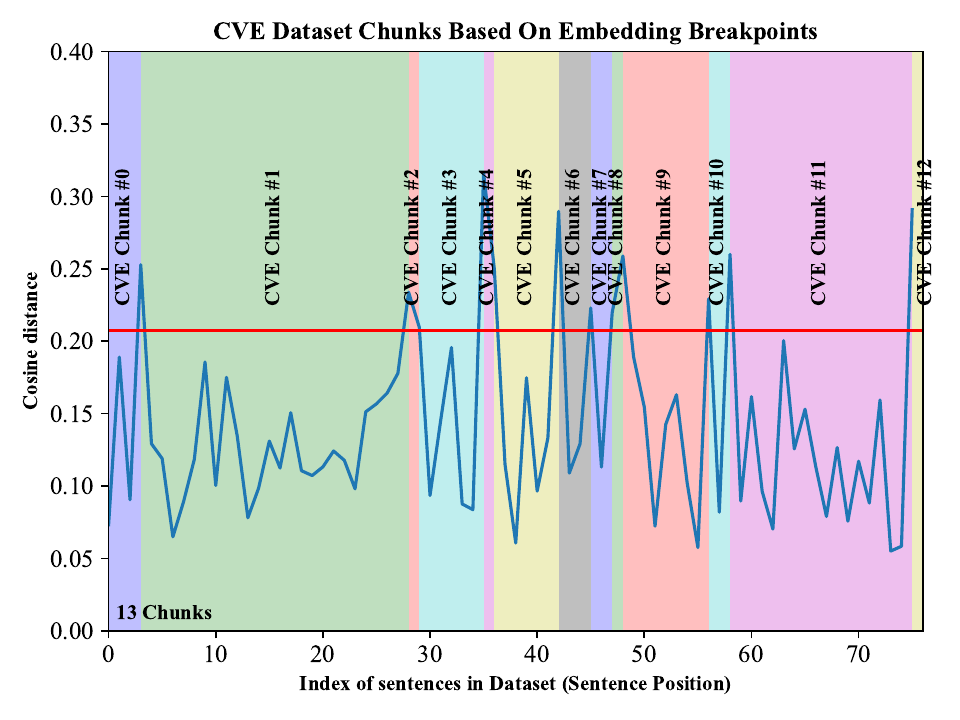}
    \caption{Distances exceeding the 85 percentile}
    \label{fig:chunks85}
\end{subfigure}
\hfill
\begin{subfigure}[b]{0.32\textwidth}
    \includegraphics[width=\textwidth]{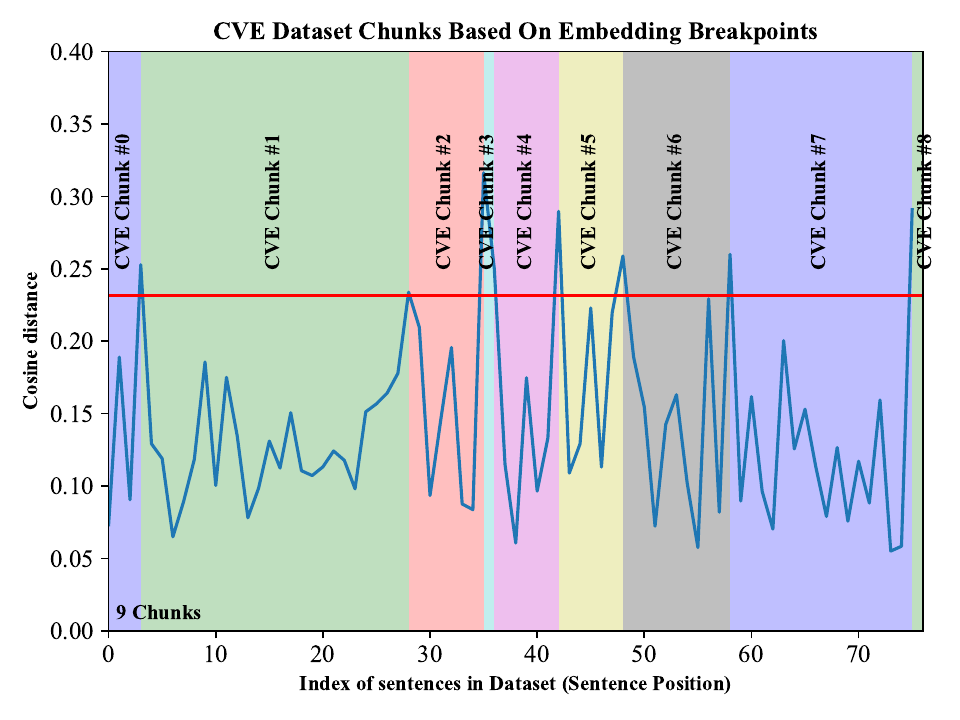}
    \caption{Distances exceeding the 90 percentile}
    \label{fig:chunks90}
\end{subfigure}
\hfill
\begin{subfigure}[b]{0.32\textwidth}
    \includegraphics[width=\textwidth]{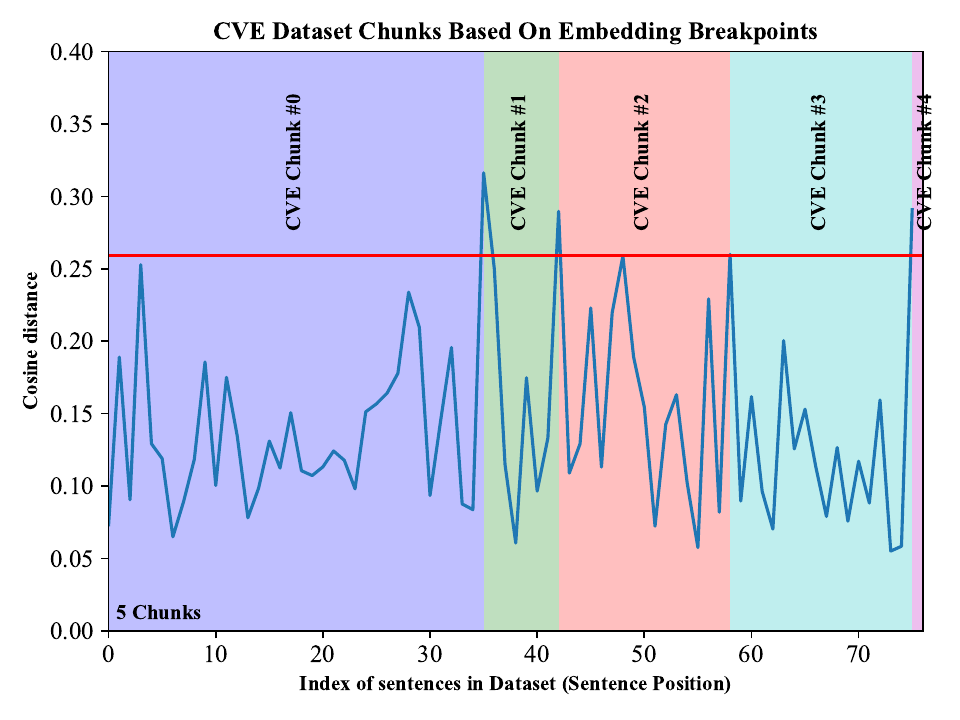}
    \caption{Distances xceeding the 95 percentile}
    \label{fig:chunks95}
\end{subfigure}
\caption{Three different values of the split point percentiles. The different colors denote the split results and the redline indicates the threshold of the split point percentiles}
\label{fig:overall-chunks}
\end{figure*}

% 在文档分块方面，我们首先使用了父文档分块技术，在此基础上增加了基于语义的文档分块技术。为了验证我们方法的有效性，说明RQ2的问题，我们对文档分块进行可视化的表示。我们选取了9个CVE文档，以CSV格式保存，在RAG的文档加载阶段，将这9个文档拼接为一个大的CSV文档。我们可以通过embedding技术进行动态规划，将文本转化为高维空间中的向量，这些向量能够反映出文本的语义内容。通过文本嵌入技术，可以捕捉到文本的深层次语义信息。当比较两段文本的嵌入向量时，可以根据它们在高维空间中的距离或者角度，来推断这两段文本在语义上的相似度或者差异。利用相似度，将语义上相似的文本自动分组在一起，形成聚类，这有助于更好地理解和组织大量的文本数据。我们基于余弦相似度进行划分，配置分割点的参数为超过距离百分位数90%的距离视为一个分割点。我们调整了分割点百分位数的值，分别比较了百分位数85% 90%和95%的情况，结果如下图所示：
In terms of document chunking, we first employed parent document chunking techniques, and then added semantic-based document chunking techniques. To validate the effectiveness of our method and address the issue of RQ2, we provide a visual representation of document chunking. We selected 9 CVE documents, saved them in CSV format, and concatenated these documents into a single large CSV document during the document loading phase of RAG. We can use embedding techniques for dynamic programming to convert the text into vectors in high-dimensional space, which can reflect the semantic content of the text. Text embedding techniques can capture the deep semantic information of the text. When comparing the embedding vectors of two pieces of text, their semantic similarity or difference can be inferred based on their distance or angle in high-dimensional space. By utilizing similarity, semantically similar texts can be automatically grouped together to form clusters, which helps in better understanding and organizing large amounts of text data. We perform the partitioning based on cosine similarity, configuring the parameter for the split point such that distances exceeding the 90 percentile are considered as split points. We adjusted the values of the split point percentiles and compared the cases of the 85 percentiles (shown in Figure \ref{fig:chunks85}), 90 percentiles(shown in Figure \ref{fig:chunks90}), and 95 percentiles (shown in Figure \ref{fig:chunks95}).

% 图中可以看出，距离百分位数为90%时恰好分为9个块，百分位数为95时分为5块，百分位数为85%时候分为13块。结合最终分块输出，这里比较理想的情况是将该参数设置为90%
% \subsection{Teacher-Guided Optimization}

\subsection{Evaluation}
% 我们使用了常见的LLM作为实验对象，其中例如GPT-4o-mini和Claude-3.5，Gemini-2.0均为最新版本，我们希望在最新的版本中测试并评价LLM是否能够解决CVE知识冲突的问题。我们将LLM分为闭源和开源两类进行实验比较。其中闭源LLM我们使用GPT-4o-mini，开源LLM我们使用Mistral和Llama3。
% 在实验结果的评价中，我们采用多种丰富的评价指标进行组合，对LLM生成内容从传统自然语言评价和与RAG相关的检索内容相关性，回答问题的准确性等方面进行评价。这样有助于更综合全面对我们的实验结果进行评价。具体评价指标表示的含义见4.3节。其中，Faith CR CP这三个评价指标专门针对RAG的方法，AR AC和AS三种方法通用于我们实验中的所有方法。最后我们也计算了不同方法的BLEU值和ROUGE-L值，作为参考。在未使用RAG的方法中，我们采用AR AC和AS评估方法，将最终输出结果与真实值进行计算得到结果。使用RAG的方法中，我们需要额外引入检索得到的原文内容，将其与输出结果、真实值构成三元组进行评估。表1全面展示了我们的评估结果，其中各个数值经过归一化后化为(0,1)之内的数字，每个结果保留两位有效数字。
In Table \ref{tab:evaluation}, We use common LLMs as experimental subjects, such as GPT-4o-mini, Claude-3.5, and Gemini-2.0, which are all the latest versions. We aim to test and evaluate whether the latest versions of LLMs can resolve CVE knowledge conflicts. We categorize the LLMs into two types: closed-source and open-source for experimental comparison. For closed-source LLMs, we use GPT-4o-mini, while for open-source LLMs, we use Mistral and Llama3.

In the evaluation of experimental results, we employed a variety of rich evaluation metrics to assess the content generated by LLMs. This included traditional natural language evaluations, relevance to RAG content, and the accuracy of answers. This approach helps to provide a more comprehensive evaluation of our experimental results. The specific meanings of the evaluation metrics are detailed in Section 4.3. Among these, the metrics Faith, CR, and CP are specifically designed for RAG methods, while AR, AC, and AS are applicable to all methods used in our experiments. Finally, we calculated the BLEU and ROUGE-L scores for different methods as a reference. For methods that did not use RAG, we employed the AR, AC, and AS evaluation methods to compute the results by comparing the final output with the true values. In methods that utilized RAG, we needed to additionally incorporate the retrieved original content, forming a triplet with the output and true values for evaluation. Table \ref{tab:evaluation} comprehensively presents our evaluation results, where each value has been normalized to fall within the range of $(0,1)$, with each result retaining two significant digits.

\subsection{Ablation Study}
% 为了综合评价我们的方法，基于上述评价指标，我们进一步设计消融实验。我们的方法分为两个阶段，因此为了充分衡量两个阶段的有效性，我们设计了消融实验对两个阶段进行单独的评估。首先我们设计了5类消融实验与我们的方法进行对比。其中，No Prompt方法和NaiveRAG方法分别作为是否引入RAG方法的基线方法作为参考。在不使用RAG的条件下，No Prompt方法为直接询问最新的LLM有关于CVE漏洞的变更信息，Prompt Only方法表示只使用提示词工程和CoT方法判断，DPO表示引入了教师指导的偏好调优过程。NaiveRAG表示最基本的RAG方案，作为对比，CRVA-TGRAG (No DPO)表示引入了第一阶段的优化RAG，但是不使用DPO的方法，最后CRVA-TGRAG方法是本文使用的方法。
% 对比来看，在引入DPO但是不使用RAG的方法中，AR值已经有了明显提升，但是AC和AS值依然与未使用DPO的方法接近，实验结果中，引入DPO的方法可以回答出一部分问题，但是效果依旧不好。我们引入并改善了RAG方法后，使用Faith CR和CP三个评价指标进行评价，相比NaiveRAG，三个评价指标的值均有明显上升。最终我们将两阶段的方法组合起来（Ours），使用上述方法进行评估，结果明显优于只使用DPO或者只使用RAG的方法。在表中我们将每一列评估的最优值和次优值，分别用加粗和加下划线方式标注出来。

To comprehensively evaluate our method, in \ref{tab:evaluation} we further designed ablation experiments based on the aforementioned evaluation metrics. Our method is divided into two stages; therefore, to adequately assess the effectiveness of both stages, we designed ablation experiments for separate evaluation. First, we designed five types of ablation experiments to compare with our method. Among these, the \textbf{No Prompt} method and \textbf{NaiveRAG} method serve as baseline methods to determine whether RAG is introduced. Under the condition of not using RAG, the \textbf{No Prompt} method directly queries the latest LLM for information on changes related to CVE vulnerabilities. The \textbf{Prompt Only} method indicates that only prompt engineering and Chain of Thought (CoT) methods are used for judgment. DPO refers to the introduction of a teacher-guided preference tuning process. \textbf{NaiveRAG} represents the most basic RAG scheme. For comparison, \textbf{CRVA-TGRAG (No DPO)} indicates the introduction of the optimized RAG from the first stage, but without using the DPO method.

In comparison, for the method that introduces DPO but does not use RAG, the AR value has shown a significant improvement, while the AC and AS values remain close to those of the method that does not use DPO. The experimental results indicate that the method incorporating DPO can answer some questions, but the performance is still unsatisfactory. After introducing and improving the RAG method, we evaluated it using the three metrics: Faith, CR, and CP. Compared to NaiveRAG, the values of the three evaluation metrics have significantly increased. Ultimately, we combined the two-stage method (\textbf{Ours}) and evaluated it using the aforementioned methods. The results clearly outperform those of methods that use only DPO or only RAG. In the table, we highlighted the optimal and suboptimal values for each column of evaluations using bold and underlined formatting, respectively.

\renewcommand{\arraystretch}{1.1}  % 1.3 倍行距
\begin{table*}[htbp]
\centering
\caption{The Evaluation Metrics Result in CVE Knowledge Conflict Resolving. We evaluate 6 methods and 7 LLMs with different parameters and size. The bold entities denote the best results in each column and the underlined ones denote the second best results. As for ROUGE-L, F denotes F-measure, P denotes Precision and R denotes Recall. The other metrics are introduced in Setion \ref{Section:Evaluation Metrics} Evaluation Metrics.}
\label{tab:evaluation}
\resizebox{\textwidth}{!}{%
\begin{tabular}{cccccccccccc}
\hline
\toprule
\multirow{3}{*}{Method} & \multirow{3}{*}{LLM} & \multicolumn{10}{c}{Metrics} \\ \cline{3-12} 
 &  & \multirow{2}{*}{Faith} & \multirow{2}{*}{CR} & \multirow{2}{*}{CP} & \multirow{2}{*}{AR} & \multirow{2}{*}{AC} & \multirow{2}{*}{AS} & \multirow{2}{*}{BLUE} & \multicolumn{3}{c}{ROUGE-L} \\ \cline{10-12} 
 &  &  &  &  &  &  &  &  & F & P & R \\ \hline
\multirow{5}{*}{No Prompt} & GPT-4o-mini & - & - & - & 0.57 & 0.32 & 0.80 & 0.03 & 0.18 & 0.12 & 0.13 \\
 & Claude-3.5-Haiku-20241022 & - & - & - & 0.47 & 0.36 & 0.80 & 0.03 & 0.15 & 0.11 & 0.11 \\
 & Gemini-2.0-Flash-Exp & - & - & - & 0.91 & 0.29 & 0.80 & 0.04 & 0.19 & 0.12 & 0.13 \\
 & Llama-3-70B & - & - & - & 0.25 & 0.35 & 0.78 & 0.03 & 0.16 & 0.10 & 0.11 \\
 & Mistral-Medium & - & - & - & 0.54 & 0.29 & 0.79 & 0.03 & 0.22 & 0.08 & 0.11 \\ \hline
\multirow{5}{*}{Prompt Only} & GPT-4o-mini & - & - & - & 0.38 & 0.35 & 0.80 & 0.03 & 0.15 & 0.11 & 0.11 \\
 & Claude-3.5-Haiku-20241022 & - & - & - & 0.01 & 0.32 & 0.78 & 0.03 & 0.10 & 0.09 & 0.08 \\
 & Gemini-2.0-Flash-Exp & - & - & - & 0.80 & 0.25 & 0.79 & 0.02 & 0.29 & 0.06 & 0.09 \\
 & Llama-3-70B & - & - & - & 0.12 & 0.29 & 0.77 & 0.02 & 0.15 & 0.08 & 0.09 \\
 & Mistral-Medium & - & - & - & 0.30 & 0.26 & 0.78 & 0.03 & 0.19 & 0.08 & 0.10 \\ \hline
\multirow{2}{*}{DPO+Prompt} & DPO-Llama3-8B & - & - & - & 0.88 & 0.26 & 0.79 & 0.02 & 0.19 & 0.09 & 0.11 \\
 & DPO-Mistral-8B & - & - & - & \underline{0.92} & 0.24 & 0.79 & 0.03 & 0.21 & 0.09 & 0.11 \\ \hline
Naive RAG & GPT-4o-mini & 0.70 & 0.01 & 0.02 & 0.01 & 0.28 & 0.78 & 0.03 & 0.10 & 0.10 & 0.09 \\ \hline
CRVA-TGRAG (No DPO) & GPT-4o-mini & 0.49 & \underline{0.46} & \underline{0.45} & \underline{0.92} & \underline{0.83} & \textbf{0.96} & \underline{0.21} & \textbf{0.59} & \textbf{0.57} & \textbf{0.51} \\ \hline

\multirow{2}{*}{CRVA-TGRAG \textbf{(Ours)}} & DPO-Llama3-8B & \underline{0.75} & \textbf{0.47} & 0.44 & 0.90 & \underline{0.83} & 0.94 & 0.20 & \underline{0.49} & 0.38 & \underline{0.38} \\ 
 & DPO-Mistral-8B & \textbf{0.76} & \underline{0.46} & \textbf{0.46} & \textbf{0.93} & \textbf{0.88} & \underline{0.95} & \textbf{0.28} & \textbf{0.59} & \underline{0.51} & \textbf{0.51} \\
 \bottomrule
\end{tabular}%
}
 \footnotesize

\end{table*}

\section{Limitation}
% 缩写问题 联网检索的准确率不一定 CVE库中的内容不一定是准确的。共性的问题：恶意SEO+正常情况下不一定准确，realtime提一句
% 我们进一步研究发现，RAG与Agent结合实现的联网在线检索方式能够获取到实施信息和数据，在一定程度上提升了检索的准确性，缓解了知识冲突问题。 \cite{}联网检索内容包括了使用搜索引擎进行全网搜索和使用指定在线数据库精确检索两个方面。然而我们注意到这两种方法存在共性问题，（1）搜索引擎受到恶意SEO的诱导， \cite{} 可能检索到内容无关的数据或者过时数据；（2）全网搜索到的内容或在线数据库的内容仍然存在更新不及时问题，检索到的内容不准确。上述问题都有可能在最终由LLM输出过程中引发知识冲突。总之，Agentic RAG 方案结合互联网搜索，进一步缓解LLM知识冲突，是未来的研究方向。
Our further research indicates that Agentic RAG with online retrieval can acquire implementation information and data, thereby enhancing retrieval accuracy and alleviating knowledge conflicts to some extent  \cite{Adaptive_RAG}. Online retrieval encompasses two aspects: conducting comprehensive searches using search engines such as Google and Bing, and performing precise searches using designated databases like NVD and CNNVD. However, we have observed common issues associated with these aspects: (1) Search engines may be influenced even induced by malicious SEO  \cite{SEO}, potentially retrieving irrelevant or outdated content; (2) Content obtained from comprehensive searches or online databases may still suffer from timeliness issues, resulting in the retrieval of inaccurate information. These problems can lead to knowledge conflicts during the final output process of the LLM. In summary, the Agentic RAG approach, which combines internet searches, represents a promising direction for future research aimed at further mitigating knowledge conflicts in LLMs.

\section{Conclusion}
In this paper, we focus on alleviating the knowledge conflict problems that LLMs have faced when dealing with CVE-IDs. We propose a innovative framework CRVA-TGRAG, which is capable of splitting, embedding, retrieving, and processing high-threat-level CVE vulnerabilities to solve the knowledge conflict. This framework integrates functionalities for data pre-processing, splitting and retriving (Stage I), and output optimization (Stage II), effectively addressing the current CVE knowledge conflict problems encountered during LLM generation and mitigating hallucination. Our framework also includes a comprehensive evaluation phase for LLM retrieval and generation results, combined with different evaluation metrics, which is making intuitive, comprehensive and systematic evaluation for the output. Our research findings and evaluations indicate that our framework achieves a high accuracy in vulnerability analysis and generation surpassing traditional direct prompting and RAG-based methods. Furthermore, our generated content offers engineers and developers detailed insights into vulnerabilities, feasible recommendations, and remediation analyses. In summary, our framework not only effectively resolves the knowledge conflict that LLMs experience when retrieving CVEs, but also significantly enhances retrieval quality, providing engineers with valuable insights and solutions to vulnerabilities.

\section*{Acknowledgments}
The information of this work was supported temporarily hidden due to anonymous submission.

\bibliographystyle{named}
\bibliography{ijcai25}

\end{document}